\documentclass[10pt,twocolumn,letterpaper]{article}

\usepackage{cvpr}
\usepackage{times}
\usepackage{epsfig}
\usepackage{graphicx}
\usepackage{amsmath}
\usepackage{amssymb}
\usepackage{color} % for \comment command

\usepackage{booktabs}

\usepackage[pagebackref=true,breaklinks=true,letterpaper=true,colorlinks,bookmarks=false]{hyperref}

\cvprfinalcopy % *** Uncomment this line for the final submission

\def\cvprPaperID{2138} % *** Enter the CVPR Paper ID here

\ifcvprfinal\fi

\newcommand{\SO}[1]{\mathbb{SO}(#1)}
\newcommand{\SE}[1]{\mathbb{SE}(#1)}
\newcommand{\R}[1]{\mathbb{R}^#1}
\newcommand{\RplusSO}[1]{\mathbb{SO}(#1)\text{ and } \mathbb{R}^#1}
\newcommand{\energy}[1]{\mathbb{E}\{#1\}}

\graphicspath{{./figures/}}

\begin{document}

%%%%%%%%% TITLE
\title{Spline Error Weighting for Robust Visual-Inertial Fusion}

\author{Hannes Ovr\'en and
Per-Erik Forss\'en\\
Link\"oping University\\
Link\"oping, Sweden\\
{\tt\small \{hannes.ovren,per-erik.forssen\}@liu.se}
}

\maketitle
%\thispagestyle{empty}

%%%%%%%%% ABSTRACT
\begin{abstract}
In this paper we derive and test a probability-based weighting that
can balance residuals of different types in spline fitting. In
contrast to previous formulations, the proposed spline error weighting
scheme also incorporates a prediction of the approximation error of
the spline fit. We demonstrate the effectiveness of the prediction in
a synthetic experiment, and apply it to visual-inertial fusion on
rolling shutter cameras. This results in a method that can estimate 3D
structure with metric scale on generic first-person videos.
We also propose a quality measure for spline fitting, that can be used to 
automatically select the knot spacing.
Experiments verify that the obtained trajectory quality corresponds
well with the requested quality. Finally, by linearly scaling the
weights, we show that the proposed spline error weighting minimizes
the estimation errors on real sequences, in terms of scale and end-point errors.
\end{abstract}

\section{Introduction}
In this paper we derive and test a probability-based weighting that
can balance residuals of different types in spline fitting.
We apply the weighting scheme to inertial-aided {\it structure
from motion} (SfM) on rolling shutter cameras, and test it on first-person
video from handheld and body-mounted cameras. In such 
videos, parts of the sequences are often difficult to use due to
excessive motion blur, or due to temporary absence of scene structure.

It is well known that inertial measurement units (IMUs) are a useful
complement to visual input. Vision provides bias-free bearings-only
measurements with high accuracy, while the IMU provides high-frequency
linear acceleration, and angular velocity measurements albeit with an
unknown bias \cite{bailey06}. Vision is thus useful to handle the IMU
bias, while the IMU can handle dropouts of visual tracking
during rapid motion, or absence of scene structure. In addition, the
IMU makes metric scale observable, also for monocular video.

\begin{figure}[t]
  \begin{center}
    \frame{\includegraphics[width=\columnwidth]{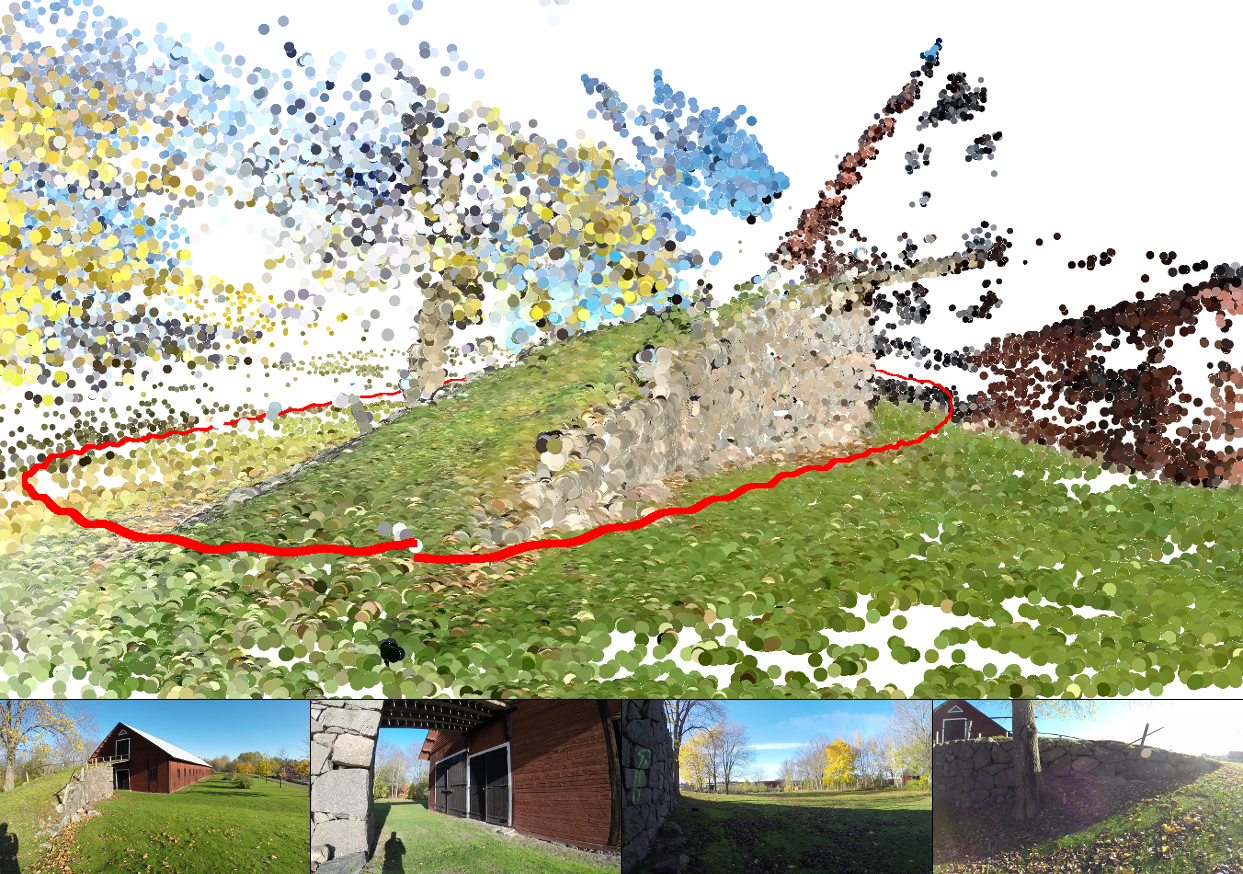}}
    \caption{Rendered model estimated on the {\bf Handheld 1} dataset. Top: model rendered using Meshlab. Bottom: Sample frames from dataset.}
    \label{fig:example_render}
  \end{center}
\end{figure}

Visual-inertial fusion using splines has traditionally balanced the
sensor modalities using inverse noise covariance
weighting \cite{lovegrove13,patron-perez15}. As we will show, this
neglects the spline approximation error, and results in an inconsistent
balancing of residuals from different modalities. 
In this paper, we propose {\it spline error weighting} (SEW), a method 
that incorporates the spline approximation error in the residual weighting.
SEW makes visual-inertial fusion robust on real sequences,
acquired with rolling shutter cameras. Figure \ref{fig:example_render}
shows an example of 3D structure and continuous camera trajectory
estimated on such a sequence.

\subsection{Related work}

Visual-inertial fusion on rolling shutter cameras has classically been
done using Extended Kalman-filters (EKF). Hanning et al.~\cite{hanning11} use an EKF to track cell-phone
orientation for the purpose of video stabilization. Li et
al.~\cite{li13} extend this to full device motion tracking, and Jia et
al.~\cite{jia13} add tracking of changes in relative pose between the
sensors and changes in linear camera intrinsics. 

Another line of work initiated in \cite{furgale12} is to define a
continuous-time estimation problem that is solved in batch, by
modelling the trajectory using temporal basis functions.
This has been done using Gaussian process (GP) regression
\cite{furgale12,anderson15}, and adapted to use hierarchical temporal
basis functions \cite{anderson14} and also a relative formulation
\cite{anderson15b} which generalizes the global shutter formulation of
Sibley et al.~\cite{sibley09}. 

The use of temporal basis functions for visual-inertial fusion can
also be made in a spline fitting framework, as was done in SplineFusion
\cite{lovegrove13,patron-perez15} for visual-inertial synchronization
and calibration. SplineFusion can be seen as a direct generalization
of bundle adjustment \cite{triggs00} to continuous-time camera paths.
A special case of this is the rolling-shutter bundle adjustment
work of Hedborg et al.~\cite{hedborg12} which linearly
interpolates adjacent camera poses to handle rolling shutter
effects. Although the formulation in \cite{lovegrove13,patron-perez15}
is aesthetically appealing, we have observed that it lacks the
equivalent of the {\it process noise} model (that defines the state
trajectory smoothness) in filtering formulations.
This makes it brittle on rolling-shutter cameras, unless the
camera motion can be well represented under the chosen knot
spacing (such as when the knot spacing is equal to the frame
distance, as in \cite{hedborg12}). A hint in this direction is
also provided in the rolling-shutter camera calibration work of Oth et
al.~\cite{oth13,furgale15}, where successive re-parametrization of the
spline (by adding knots in intervals with large residuals) was
required to attain an accurate calibration.
In this paper we amend the SplineFusion approach
\cite{lovegrove13,patron-perez15} by making the trajectory
approximation error explicit. This makes the approach more generally applicable
in situations where the camera motion is not perfectly smooth, without
requiring successive re-parametrization.

Also related to bundle adjustment is the factor graph approach. Forster et
al.~\cite{forster15} study visual-inertial fusion with
preintegration of IMU measurements between keyframes, with a global
shutter camera model.

The method we propose is generally applicable to many spline-fitting
problems, as it provides a statistically optimal way to balance
residuals of different types. We apply the method to structure from
motion on first-person videos. This problem has previously been 
studied for the purpose of geometry based video stabilization,
which was intended for high speed-up playback of the video \cite{kopf14}.
In such situations, a {\it proxy geometry} is sufficient, as
significant geometric artifacts tend not to be noticed during high
speed playback. We instead aim for accuracy of the reconstructed 3D
models, and thus employ a continuous-time camera motion model, that
can accurately model the rolling shutter present on most video cameras.

\subsection{Contributions}

\begin{itemize}
\item{} We derive expressions for {\it spline error weighting} (SEW) and
  apply these to the {\it continuous-time structure from motion}
  (CT-SfM) problem. We verify experimentally that the proposed
  weighting produces more accurate and stable trajectories than the
  previously used inverse noise covariance weighting.
\item{} We propose a criterion to automatically set a suitable knot
  spacing, based on allowed approximation error. Previously knot
  spacing has been set heuristically, or iteratively using
  re-optimization. We also verify experimentally that the obtained
  approximation error is similar to the requested.
\end{itemize}

\subsection{Notation}

We denote signals by lower case letters indexed by a time variable,
\eg $x(t)$, and their corresponding Fourier transforms in capitals
indexed by a frequency variable, \eg $X(f)$. Bold lower case letters,
\eg ${\bf x}$, denote vectors of signal values, and the corresponding
Discrete Fourier Transform is denoted by bold capital letters, \eg ${\bf
  X}$. An estimate of a value, $q$, is denoted by $\hat{q}$.

\section{Spline Error Weighting}
\label{sec:model_errors}

Energy-based optimization is a popular tool in model fitting. It
involves defining an energy function $J(\mathbf{\Theta})$ of the model
parameters $\mathbf{\Theta}$, with terms for
measurement residuals. Measurements from several different
modalities are balanced by introducing modality weights $\gamma_i$ :
\begin{align}
J(\mathbf{\Theta}) &= \gamma_x\sum_k \|x_k-\hat{x}_k(\mathbf{\Theta})\|^2 \nonumber \\
& +\gamma_y\sum_l \|y_l-\hat{y}_l(\mathbf{\Theta})\|^2 \nonumber \\
& +\gamma_z\sum_m \|z_m-\hat{z}_m(\mathbf{\Theta})\|^2\,.
\label{eq:energy}
\end{align}
Here $x, y,$ and $z$ are three measurement modalities, that are
balanced by the weights $\gamma_x, \gamma_y, \gamma_z$.

It is well known that the minimization of \eqref{eq:energy} can be
expressed as the maximisation of a probability, by exponentiating and
changing the sign. This results in:
\begin{equation}
p(\mathbf{\Theta})=\prod_kp_k(x_k|\mathbf{\Theta})
\prod_lp_l(y_l|\mathbf{\Theta})
\prod_mp_m(z_m|\mathbf{\Theta})\,.
\end{equation}
For the common case of normally distributed measurement residuals, we
have:
\begin{equation}
p_k(x_k|\mathbf{\Theta})\propto e^{\displaystyle -(x_k-\hat{x}_k(\mathbf{\Theta}))^2/2\sigma_x^2}\,,
\end{equation}
where $\sigma_x^2=1/(2\gamma_x)$ is the variance of the residual
distribution.

In the context of splines, $\boldsymbol{\Theta}$ is a coefficient
vector, and depending on the knot density, the predictions,
$\hat{x}(t|\boldsymbol{\Theta})$ will cause an {\it approximation
  error}, $e(t)$, if the spline is too 
smooth to predict the measurements during rapid changes. We thus have
a residual model:
\begin{equation}
r(t)=x(t)-\hat{x}(t|\boldsymbol{\Theta})=n(t)+e(t)\,,
\end{equation}
where $n(t)$ is the measurement noise. In the SplineFusion approach
\cite{patron-perez15}, the variances that balance the optimization are
set to the measurement noise variance, thereby neglecting $e(t)$. We
will now derive a more accurate residual variance, based on signal
frequency content.

\subsection{Spline fitting in the frequency domain}
\label{sec:frequency_response}

Spline fitting can be characterized in terms of a frequency response
function, $H(f)$, see \cite{unser1993,mihajlovic1999}. 
In this formulation, a signal $x(t)$ with the Discrete Fourier Transform (DFT)
$X(f)$ will have the frequency content $(H\cdot X)(f)$ after spline
fitting. In \cite{mihajlovic1999}, closed form expressions of $H(f)$ are
provided for B-splines of varying orders. 
By denoting the DFT of the frequency response function by the vector
${\bf H}$, and the DFT of the signal by ${\bf X}$, we can express the
error introduced by the spline fit as:
\begin{equation}
{\bf E}=(1-{\bf H})\cdot {\bf X}\,.
\label{eq:approximation_error}
\end{equation}
We now define the inverse DFT operator as an $N\times N$ matrix ${\bf
  M}$ with elements $M_{kn}=\frac{1}{\sqrt{N}}e^{{\bf i}2\pi kn/N}$,
for which ${\bf M}^T{\bf M}={\bf I}$. 
Now we can compute the error signal $e(t)$, as ${\bf e}={\bf M}{\bf E}$, and its
variance $\hat{\sigma}_e^2$ as:
\begin{equation}
\hat{\sigma}_e^2=\energy{{\bf E}}/N=\energy{(1-{\bf H})\cdot{\bf X}}/N\,,
\label{eq:approximation_variance}
\end{equation}
where $\energy{{\bf X}}=\|{\bf X}\|^2$. The variance expression above
follows directly from the Parseval theorem, as is easy to show:
\begin{equation}
N\hat{\sigma}_e^2=\sum_{t=1}^N e(t)^2={\bf e}^T{\bf e}={\bf
  E}^T{\bf M}^T{\bf M}{\bf E}={\bf E}^T{\bf E}
\end{equation}
Here we have used the fact that $H(0)=1$ for all spline fits (see \cite{mihajlovic1999}),
and thus $e(t)$ is a zero-mean signal.

To obtain the final residual error prediction, our estimate of the
approximation variance in \eqref{eq:approximation_variance}, should
be added to the noise variance that was used in \cite{patron-perez15}.
However, the spline fit splits the noise in two parts ${\bf N}=(1-{\bf
  H}) \cdot {\bf N}+{\bf H} \cdot {\bf N}$. Now, as
  \eqref{eq:approximation_variance} is estimated using the actual, 
noisy input signal ${\bf X}={\bf X}_0+{\bf N}$ it will already
incorporate the part of the noise ${\bf N}$ that the spline filters out:
\begin{equation}
{\bf E}=(1-{\bf H})\cdot {\bf X}=(1-{\bf H}) \cdot {\bf X}_0+(1-{\bf H}) \cdot {\bf N}\,.
\end{equation}
We should thus add only the part of the noise that was kept. We denote
this {\it filtered} noise term by ${\bf F}={\bf H} \cdot {\bf N}$, and its
variance by $\hat{\sigma}_f^2$. We can now state the final expression of the
residual noise prediction:
\begin{equation}
\hat{\sigma}_r^2=\hat{\sigma}_e^2+\hat{\sigma}_f^2\,.
\label{eq:residual_error}
\end{equation}
For white noise, the filtered noise variance can be estimated from the measurement 
noise $\sigma_n$ and ${\bf H}$ as:
\begin{equation}
\hat{\sigma}_f^2 = \sigma_n^2\energy{{\bf H}}/N\,.
\end{equation}

The final weight to use for each residual modality (see
\eqref{eq:energy}) is the inverse of its predicted residual error variance:
\begin{equation}
\gamma = \frac{1}{\hat{\sigma}_r^2}\,.
\label{eq:weight}
\end{equation}

\begin{figure}[t]
\includegraphics{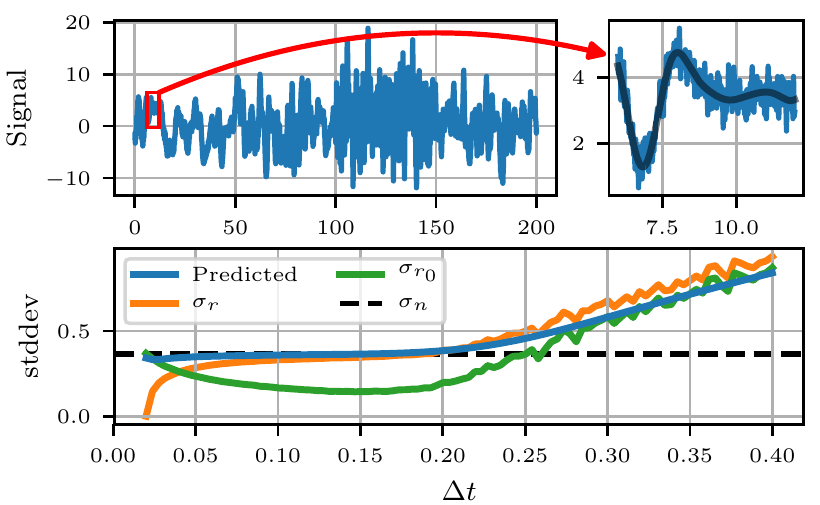}
\caption{Top: 50Hz test signal and noise (right subplot is a
  detail). Bottom: standard deviations as functions of knot
  spacing. $\sigma_r$ is the empirical residual standard deviation, 
  $\sigma_n$ is the noise standard deviation, which is used in
  \cite{patron-perez15} to predict $\sigma_r$, {\it Predicted} is the proposed residual
  noise prediction. $\sigma_{r_0}$ is the residual with respect to the noise-free signal $x_0(t)$.}
\label{fig:splinefit}
\end{figure}

\subsection{A simple 1D illustration}
\label{sec:simple}

In figure \ref{fig:splinefit} we illustrate a simple experiment that
demonstrates the behaviour of our proposed residual error prediction \eqref{eq:residual_error}. 
In figure \ref{fig:splinefit} top left, we show a test signal,
$x(t)$, which is the sum of a true signal, $x_0(t)$, and white
Gaussian noise $n(t)$ with variance $\sigma_n^2$. The true
signal has been generated by filtering white noise to produce a range
of different frequencies and amplitudes. In figure
\ref{fig:splinefit} top right, we show a detail of the signal, where
the added noise is visible.

We now apply a least-squares spline fit to the signal $x(t)$, to
obtain the spline $\hat{x}(t)$, with control points $\boldsymbol{\Theta} = (\theta_1,\ldots,\theta_K)^T$
and basis functions $B(t)$:
\begin{equation}
\hat{x}(t|\boldsymbol{\Theta}) = \sum_{k=1}^K \theta_kB(t-k\Delta t)\,.
\end{equation}
This is repeated for a range of knot spacings, $\Delta t$, each
resulting in a different residual $r(t)=x(t)-\hat{x}(t)$. The residual
standard deviation $\sigma_r$ is plotted in figure \ref{fig:splinefit}, bottom.
We make the same plot for the residual $r_0(t)=x_0(t) - \hat{x}(t)$ which measures
the error compared to the true signal. The resulting $\sigma_{r_0}$ curve has
a minimum at approximately $\Delta t=0.15$, which is thus the optimal knot
spacing. The fact that the actual residual $\sigma_r$ decreases for
knot spacings below this value thus indicates overfitting. From the
experiment, we can also see that the implicit assumption made in
\cite{patron-perez15} that the noise standard deviation $\sigma_n$ can 
predict $\sigma_r$ is reasonable for knot spacings at or below the
optimal value. However, for larger knot spacings (at the right side of
the plot) this assumption becomes increasingly inaccurate.

\subsection{Selecting the knot spacing}
\label{sec:knot_spacing}

Instead of deciding on a knot spacing explicitly, a more convenient
design criterion is the amount of approximation error introduced by
the spline fit. To select a suitable knot spacing,
$\Delta t$, we thus first decide on a {\it quality value}, $\hat q \in
(0, 1]$, that corresponds to the fraction of signal energy we want the
approximation to retain. For a given signal, $x(t)$, with the DFT,
${\bf X}$, we define the quality value as the ratio between the
energy, before and after spline fitting:
\begin{align}
  q(\Delta t) = \frac{\energy{{\bf H}(\Delta t) \cdot {\bf X}}}{\energy{{\bf X}}}
  \label{eq:quality}
\end{align}

Here ${\bf H}(\Delta t)$ is the scaled frequency response of the
spline approximation, see section \ref{sec:frequency_response}. By
constraining $H(f)$ to preserve the DC-component after a change of
variables, \ie, $H(0)=1$, the change of variables simplifies to a
scaling of the original frequency response, $H(f,\Delta t)=H(\Delta t f)$\,.

To find a suitable knot spacing for the signal, we search for the
largest knot spacing $\Delta t$ for which $q(\Delta t) \geq \hat q$.
We do this by starting at the maximum allowed knot spacing, and then
decrease $\Delta t$ until $q(\Delta t) \geq \hat q$.
At this point we use the bounded root search method by Brent  \cite{brent73}
to find the exact point, $\Delta t$, where $q(\Delta t) = \hat q$.

\section{Visual-inertial fusion}
\label{sec:spline_fusion}

We will use the residual error prediction introduced in section \ref{sec:model_errors}
 to balance visual-inertial fusion on rolling shutter cameras. Our
 formulation is largely the same as in the {\it SplineFusion} method
 \cite{patron-perez15}. 
In addition to the improved balancing, we modify the SplineFusion
method, as follows: (1) we interpolate in
$\RplusSO3$ instead of in $\SE3$, (2) we use a 
rolling shutter reprojection method based on observation time\footnote{In \eqref{eq:cost_function} we set $t_{k,m} = t_{m} + (r \cdot v_{k,m}) / N_v$, where $v_{k,m}$ is the observation row coordinate, $r$ is the readout time, $N_v$ is number of rows, and $t_m$ is the start time of frame $m$.}
instead of performing Newton-optimization,
and (3) we add a robust error norm for image residuals.

Given IMU measurements $\{\boldsymbol{\omega}_n\}_1^N,\{{\bf a}_l\}_1^L$, and tracked image points, the objective is to
estimate the trajectory ${\bf T}(t)$, and the 3D landmarks that best
explain the data.
This is done by minimizing a cost function $J(\boldsymbol{\theta},\boldsymbol{\rho})$
where $\boldsymbol{\theta}$ and $\boldsymbol{\rho}$ are trajectory and landmark
parameters, respectively.
We parameterize landmarks using the inverse depth relative to a reference observation,
while the trajectory parameters are simply the spline control points
for the cubic B-splines.
We define our cost function as
\interdisplaylinepenalty=10000
\begin{align}
  J(\boldsymbol{\theta},\boldsymbol{\rho}) =& \sum_{k,m} \phi({\bf x}_{k,m}-\pi({\bf
  x}_{k,0},{\bf T}(t_{k,m}) {\bf T}(t_{k,0})^{-1},\rho_k)) \notag\\
+&\sum_n||\boldsymbol{\omega}_n-\nabla_{\omega}{\bf T}(t_n)||^2_{{\bf W}_g}\label{eq:cost_function}\\
+&\sum_l||{\bf a}_l-\nabla^2_{a}{\bf T}(t_l)||^2_{{\bf W}_a}\,. \notag
\end{align}
Here, $\phi$ is a robust error norm. Each landmark observation ${\bf x}_{k,m}$ belongs
to a track $\left\{{\bf x}_{k,m}\right\}_{m=0}^M$, and has an
associated inverse depth $\rho_k$. The function $\pi(\cdot)$
reprojects the first landmark observation into subsequent
frames using the trajectory and the inverse depth.
The norm weight matrices ${\bf W}_g$ and ${\bf W}_a$ are in general
matrices, but in our experiments they are assumed to be isotropic,
which results in 
\begin{equation}
{\bf W}_g={\bf I}\frac{1}{\hat{\sigma}_{r,g}^2}\,,\text{ and } {\bf W}_a={\bf
  I}\frac{1}{\hat{\sigma}_{r,a}^2}\,,
\label{eq:modality_weights}
\end{equation}
 where $\hat{\sigma}_{r,g}^2$ and $\hat{\sigma}_{r,a}^2$ are the
 predicted residual variances for the two modalities, see \eqref{eq:weight}.

The operators $\nabla_\omega$ and $\nabla^2_a$ in
\eqref{eq:cost_function} represent inertial sensor models which
predict gyroscope and accelerometer readings given the trajectory model ${\bf T}(t)$,
using analytic differentiation.
The inertial sensor models should account for at least biases in the 
accelerometer and gyroscope, but could also involve \eg
axis misalignment. 

The robust error norm $\phi$ is required for the image residuals since we expect
the image measurements to contain outliers that, unless handled, will result
in a biased result. We use the {\it Huber} error norm with a cut-off
parameter $c=2$. The gyroscope and accelerometer measurements do not
contain outliers, and thus no robust error norm is required here.

\subsection{Quality measurement domains}
\label{sec:quality}

In order to apply the method in section \ref{sec:model_errors}
on a spline defining a camera trajectory, we need to generalize
estimation to vector-valued signals, and also to apply it in 
measurement domains that correspond to derivatives of the sought trajectory. 

The gyroscope senses angular velocity, which is the derivative of
the $\SO3$ part of the sought trajectory. From the derivative theorem of
B-splines \cite{unser99}, we know that the derivative of a spline is
a spline of degree one less, with knots shifted
by half the knot spacing. Thus, as the knot spacing in the derivative domain is
the same, we can impose a quality value $\hat{q}_\text{g}$
on the gyro signal to obtain a knot spacing $\Delta t_\text{g}$ for the
$\SO3$ spline.

In order to apply \eqref{eq:quality} to the gyroscope signal,
$\boldsymbol{\omega}(t)$, we need to convert it to a
1D spectrum $X_g(f)$. This is done by first applying the DFT along the
temporal axis. We then compute a single scalar for each frequency
component using the scaled $L_2$-norm:
\begin{equation}
X_g(f)=\sqrt{\frac{1}{3}}\|\boldsymbol{\Omega}(f)\|\,\;\text{where}\;\;
\boldsymbol{\Omega}(f)=\begin{bmatrix}
\Omega_x(f)\\
\Omega_y(f)\\
\Omega_z(f)
\end{bmatrix}\,.
\label{eq:gyro_ref_signal} 
\end{equation}

We also set $X_g(0)=0$ to avoid that a large DC component dominates
\eqref{eq:quality}. Conceptually this also makes sense as a
spline depends only on the shape of the measurements, and not the
choice of origin.

To find a knot spacing for the $\R3$-spline, the connection is not as
straightforward as for the $\SO3$ part of the trajectory. Here we
employ the IMU data from the accelerometer, and use the same formulation as in
\eqref{eq:gyro_ref_signal} to compute a 1D spectrum $X_a(f)$, again with
zero DC, $X_a(0)=0$.

In contrast to the gyroscope, however, the accelerometer measurements
are induced by changes in both $\R3$ and $\SO3$: the linear acceleration, and the
current orientation of the gravity vector. Because of the
gravitational part, the measurements are always in the order of $1g$, the standard gravity, which is very large compared to most linear accelerations.
For a camera with small rotation in pitch and roll, most of the gravity component
will end up in the DC component, and thus not influence the
result. For large rotations, however, the accelerometer will
have an energy distribution that is different from the position
sequence of the final spline.

In the experiment section, we evaluate the effectiveness of the
quality estimates for both gyroscope and accelerometer.

\section{Experiments}
Our experiments are mainly designed to verify the spline error
weighting method (SEW), presented in sections \ref{sec:model_errors}
and \ref{sec:spline_fusion}.
However, they also hint at some of the benefits of incorporating inertial measurements 
in SfM: metric scale, handling of visual dropout, and reduced need for
initialization.
The experiments are all based on real data, 
complementing the synthetic example already presented in section \ref{sec:simple}.

\subsection{Datasets}
We recorded datasets for two use-cases: {\bf Bodycam} and {\bf Handheld}.
All sets were recorded outdoors using a \emph{GoPro Hero 3+ Black} camera with
the video mode set to 1080p at 30Hz.
For the bodycam datasets the camera was strapped securely to the wearer's chest
using a harness, pointing straight forward.
For the handheld datasets, the camera motion was less constrained but was mostly
pointed forwards and slightly sideways.

To log IMU data we used a custom-made IMU logger based on the
\emph{InvenSense MPU-9250} 9-axis IMU. Data was captured at 1000 Hz,
but was downsampled to 300 Hz for the experiments. 

The camera was calibrated using the FOV model in \cite{devernay01}.
An initial estimate of the time offset between the camera and IMU, as well
as their relative orientation, was found using the software package \emph{Crisp} \cite{ovren15}.
Since Crisp does not handle accelerometer biases we refined the 
calibration using the same spline-based SfM pipeline as used
in the experiments. Calibration was refined on a short time interval
with IMU biases, time offset, and relative orientation as additional
parameters to estimate.

When recording each dataset, we used a tripod with an indicator bar to
ensure that each dataset ended with the camera returned to its
starting position ($\pm 2$ cm). This enables us to use endpoint error (EPE)
as a metric in the experiments, $\text{EPE}=\|{\bf p}(t_0)-{\bf
p}(t_\text{end})\|$, where ${\bf p}(t)$ is the positional spline. To
gauge the scale error after reconstruction, we measured a few
distances in each scene using a measuring tape.

\begin{figure}[t]
  \begin{center}
    \frame{\includegraphics[width=\columnwidth]{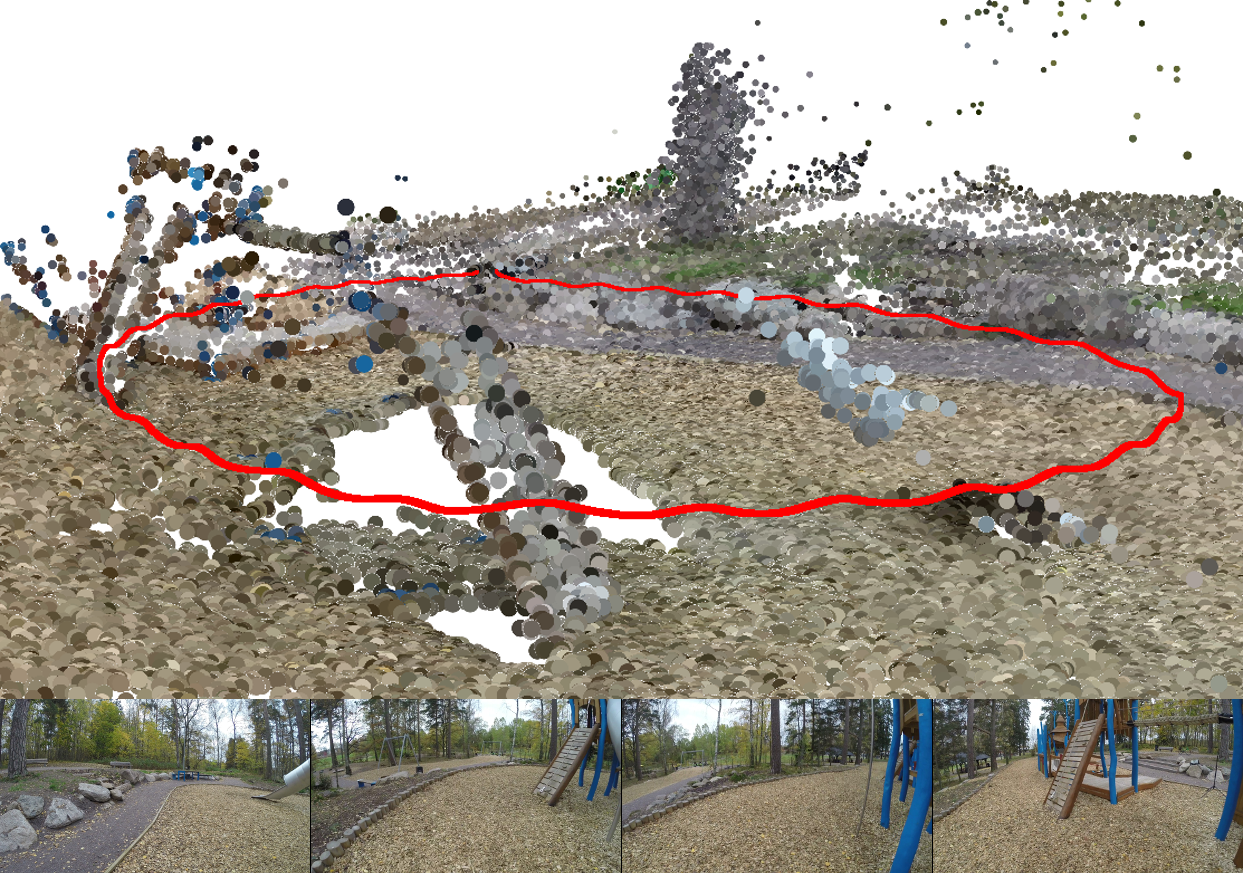}}
    \caption{Render of model estimated on {\bf Bodycam 1}
      dataset. Top: model rendered using Meshlab. Bottom: Sample
      frames from dataset.}
    \label{fig:render_playground1_montage}
  \end{center}
\end{figure}

\begin{figure}[t]
  \begin{center}
    \frame{\includegraphics[width=\columnwidth]{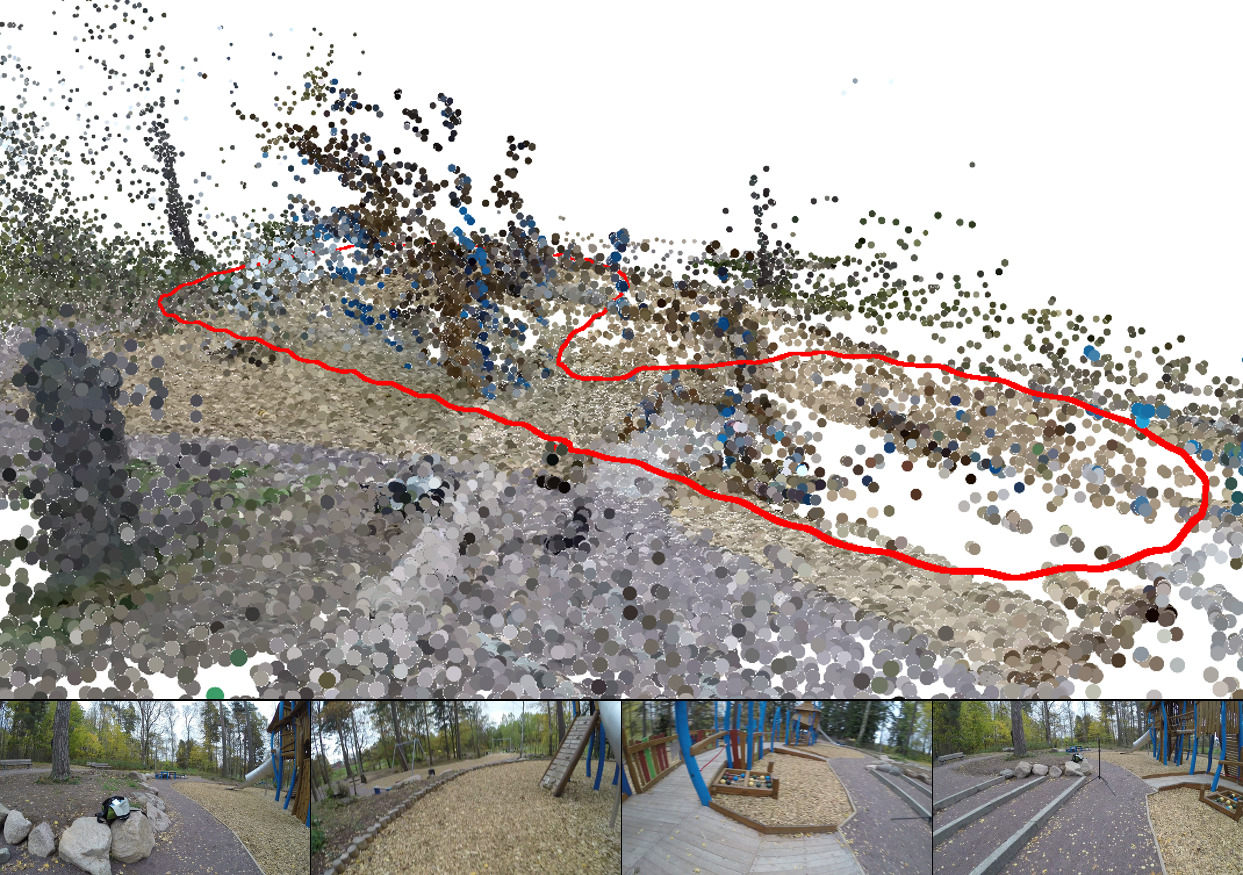}}
    \caption{Render of model estimated on {\bf Bodycam 2}
      dataset. Top: model rendered using Meshlab. Bottom: Sample
      frames from dataset.}
    \label{fig:render_playground3_montage}
  \end{center}
\end{figure}

\begin{figure}[t]
  \begin{center}
    \frame{\includegraphics[width=\columnwidth]{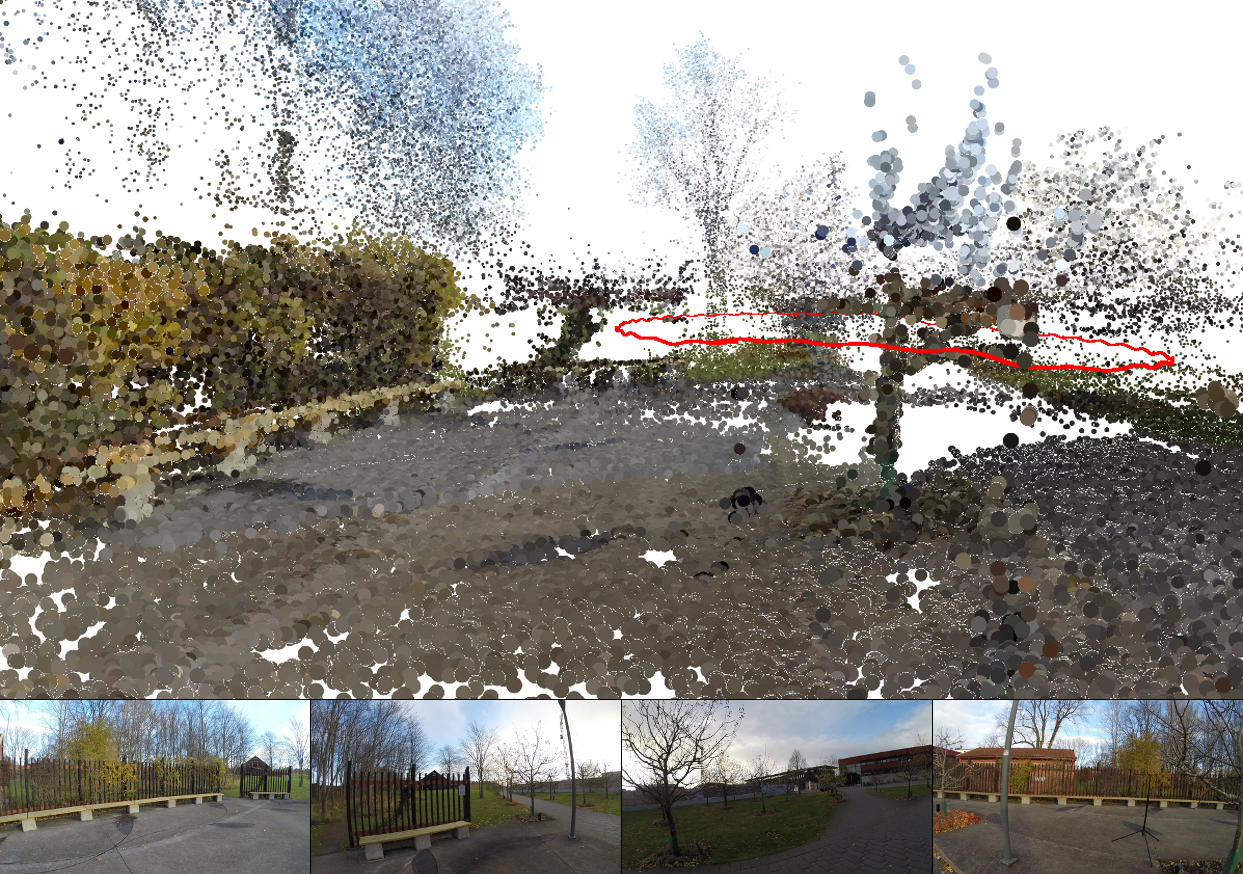}}
    \caption{Render of model estimated on {\bf Handheld 2} dataset. Top: model rendered using Meshlab. Bottom: Sample frames from dataset.}
    \label{fig:render_apple3_montage}
  \end{center}
\end{figure}

Example frames and final reconstructions of the datasets can be seen in figures \ref{fig:example_render}, \ref{fig:render_playground1_montage}, \ref{fig:render_playground3_montage}, and \ref{fig:render_apple3_montage}.
To improve visualization the example figures have been densified by triangulating additional landmarks using the estimated trajectory.

\subsection{Structure from motion pipeline}
To not distract from the presented theory, we opted for a very simple
structure from motion pipeline that can serve as a
baseline to build on. 

We generate visual observations by detecting FAST keypoints \cite{rosten10} with a regular
spatial distribution enforced using ANMS \cite{gauglitz2011}.
These are then tracked in subsequent frames using the
OpenCV KLT-tracker \cite{bouguet00}.
For added robustness, we performed backtracking and discarded tracks which
did not return to within $0.5$ pixels of its starting point.
Using tracking instead of feature matching means that landmarks that
are detected more than once will be tracked multiple times by the system.
In addition, the visual data does not contain any explicit loop closures,
which in turn means that correct weighting of the IMU residuals become more important
to successfully reconstruct the trajectory.

The trajectory splines are initialized with their respective chosen knot spacing,
such that they cover all frames in the data.
The spline control points are initialized to $\mathbf{0} \in \R3$ and
${\bf I} \in \SO3$ respectively.
All landmarks are placed at infinity, with inverse depths $\rho = 0$.
Note that we are \emph{not} using any initialization scheme like
essential matrix estimation and triangulation to give the system a
reasonable starting point.

The pipeline consists of four phases:
\begin{description}
  \item[Initial]\vspace{-1ex} Rough reconstruction using keyframes.
  \item[Cleanup ($\times 2$)] \vspace{-1ex} Reoptimize using only landmarks with a 
  mean reprojection error below a threshold.
  \item[Final] \vspace{-1ex} Optimize over all frames.
\end{description}

We now describe the phases in more detail.
To start the reconstruction we first select a set of \emph{keyframes}.
Starting with the first frame we insert a new keyframe whenever the
number of tracked landmarks from the previous keyframe drops below $75\%$.
To avoid generating too many keyframes we add the constraint that the
distance between two keyframes must be at least 6 frames.

The starting set of observations are then chosen from the keyframes
using ANMS \cite{gauglitz2011}, weighted by track lengths,
such that each keyframe provides at most $n=100$ observations.
After the initial phase is completed, we do two cleanup phases.
During a cleanup phase we first determine the mean reprojection error for all landmarks that
are visible in the keyframes.
We then pick a new set of observations (at most $n$ per frame) from
landmarks with mean reprojection errors below a threshold.
These thresholds are set to $8$ and $5$ pixels, respectively.

For the final phase we select the landmarks with a mean reprojection error below $3$ pixels, and add
observations for all frames.

\subsection{Prediction of quality}
\begin{figure}
  \begin{center}
    \includegraphics[width=\columnwidth]{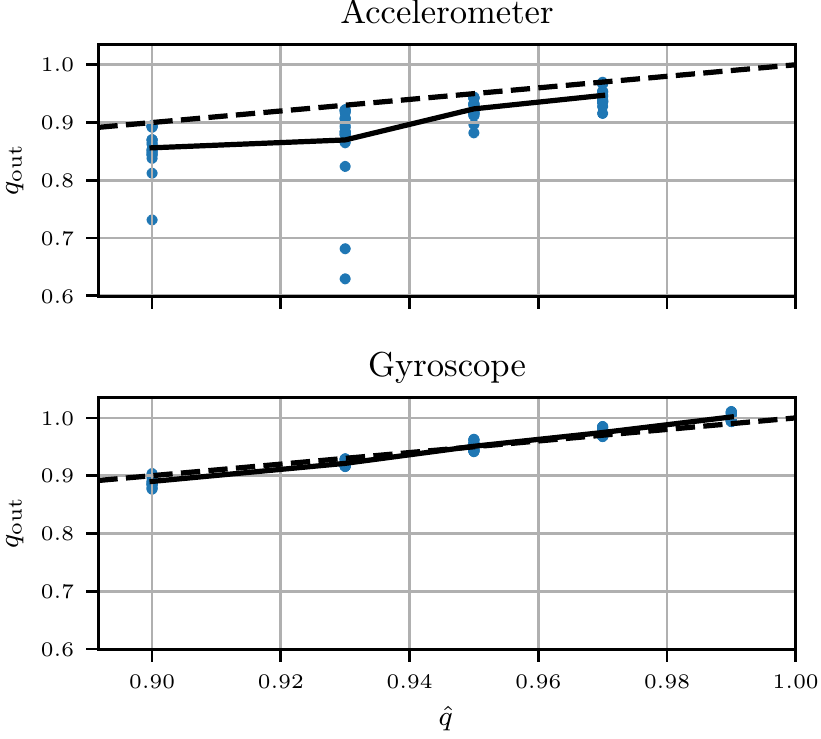}
    \caption{Actual quality ($q_\text{out}$) as a function of
      requested quality ($\hat{q}$) for both accelerometer and gyroscope.
    The blue markers are the individual data points, and the solid line their mean.
    The dashed line shows the ideal $q_\text{out} = \hat{q}$.}
    \label{fig:exp_quality}
  \end{center}
\end{figure}

We now test how well the requested quality $\hat{q}$ corresponds to
the obtained quality after spline fit.
For each dataset we performed reconstructions with a range of different quality values
for the accelerometer and gyroscope.
The requested quality value then determines the knot spacing, and IMU weights
used by the optimizer according to the theory presented in sections
\ref{sec:model_errors} and \ref{sec:spline_fusion}.
All reconstructions for a dataset were initialized with the same set of keyframes and initial observations.

As we can see in figure \ref{fig:exp_quality} the obtained quality for the gyroscope
corresponds well with the requested value.
For the accelerometer however, the obtained quality is consistently slightly lower.
The outliers in the accelerometer plot correspond to the lowest gyroscope quality values.
Since the accelerometer measurements depend also on the orientation estimate, it is
expected that a reduction in gyroscope quality would influence the
accelerometer quality as well, see section \ref{sec:quality}.

\subsection{Importance of correct weighting}

\begin{figure}
  \begin{center}
    \includegraphics[width=\columnwidth]{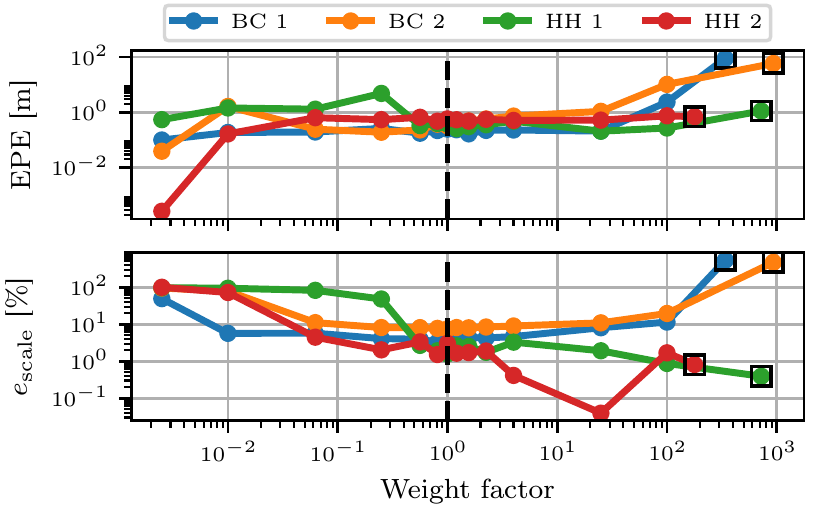}
    \caption{Endpoint error and scale error as functions of the IMU residual weights.
    The weight factor was multiplied with the base weights found using \eqref{eq:modality_weights}.
    The dashed line shows the weights selected by SEW.
    Measurements surrounded by a box represent the weighting used in SplineFusion. BC: Bodycam, HH: Handheld.}
    \label{fig:exp_weights}
  \end{center}
\end{figure}

To investigate how well our method sets the IMU residual weights,
we made several reconstructions using the same knot spacing but different weights.
The knot spacing and base weights for the IMU residuals were first
selected using the theory described in sections \ref{sec:model_errors} and \ref{sec:spline_fusion}.
We then performed multiple reconstructions with the IMU weights scaled
by a common factor.
All reconstructions for a given dataset used the same initial set of
keyframes and observations.
In figure \ref{fig:exp_weights} we plot the endpoint error and scale error as functions of this factor.
The scale error is defined as $e_\text{scale} = |l_\text{true} - \hat l|/l_\text{true}$,
where $l_\text{true}$ is the true length as measured in the real world, and $\hat l$ is a length
that is triangulated, from manually selected points, using the reconstructed trajectory.

Figure \ref{fig:exp_weights} shows that SEW produces weights which are
in the optimal band where both errors are low.
In contrast, the inverse noise covariance weights used by SplineFusion
(the right-most data points) are consistently a bad choice.

\subsection{Visual dropout}
\begin{figure}
  \begin{center}
    \includegraphics[width=\columnwidth]{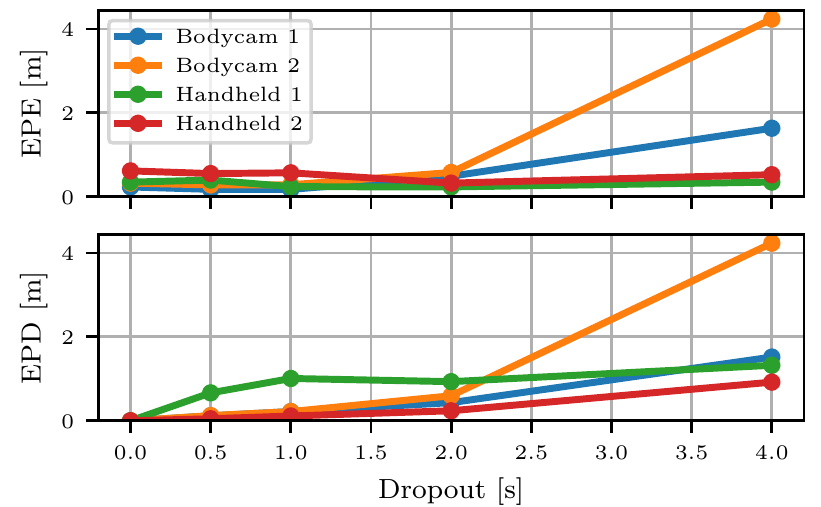}
    \caption{Endpoint error and endpoint distortion as functions of dropout time.
    The dropout is the number of seconds for which no visual observations were available.}
    \label{fig:exp_dropout}
  \end{center}
\end{figure}

One of the benefits of including IMU data in structure from motion is that we can handle
short intervals of missing visual observations.
These could be due to changes in lighting, or because the observed scenery has no
visual features (white wall, clear sky, \etc).

To investigate visual dropout in a controlled manner we simulate dropouts by removing
the last $n$ frames in each dataset.
The trajectory estimate in this part will then depend on IMU measurements only.
In figure \ref{fig:exp_dropout} we show the endpoint error and \emph{endpoint distortion} (EPD)
for different lengths of dropout.
The endpoint distortion is the distance from the estimated endpoint
under dropout to the endpoint without dropout. It thus directly
measures the drift caused by the dropout.

As expected the errors increase with the dropout time, but are quite small for dropout times below a second.
The dropout error is sensitive to the IMU/camera calibration, and especially 
to correct modeling of the IMU biases. Given that these are
accurately estimated, the IMU provides excellent support in case of
brief visual dropout.

\subsection{Comparison to SplineFusion}
Finally we make an explicit comparison with SplineFusion as described in \cite{patron-perez15},
but with the three modifications stated in section \ref{sec:spline_fusion}.
The only difference between SEW and SplineFusion in the following comparison is 
with regards to knot spacing and IMU residual weights:
For SplineFusion we set both spline knot spacings to $\Delta t = 0.1$ seconds, and IMU residual weights to the inverse of the
measurement noise covariances, as was done in \cite{patron-perez15}.
For SEW (the proposed method) the knot spacings and residual
weights were chosen using quality values $q_\text{acc}=0.97$ and
$q_\text{gyro}=0.99$. This resulted in knot spacings averaging around $0.04$ seconds.

\begin{figure}
  \begin{center}
    \includegraphics{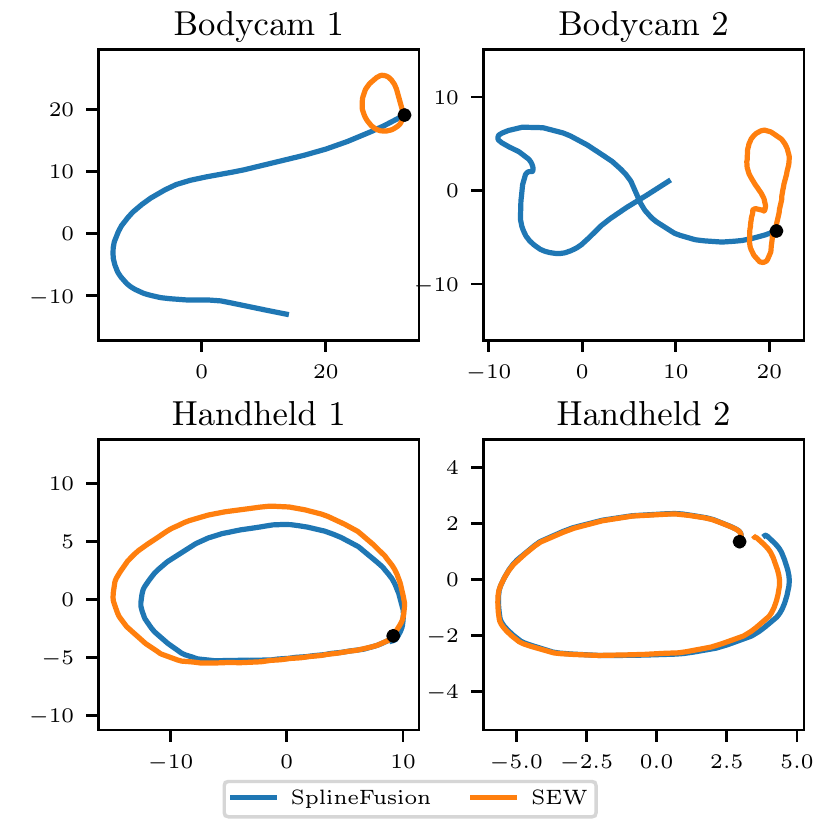}
    \caption{Comparison of obtained trajectories for SplineFusion and SEW. 
    Trajectories have been aligned to share the starting point (black dot).
    All trajectories should ideally start and end at the same point.}
    \label{fig:sf_comp_multi}
  \end{center}
\end{figure}

In figure \ref{fig:sf_comp_multi} we show the reconstructed trajectories projected onto the XY-plane,
and table \ref{tab:sf_comp} shows the corresponding endpoint and scale errors.
A perfect reconstruction should have the trajectories start and end at exactly the same point,
with zero endpoint error.
It is clear that the SplineFusion settings work reasonably well in the
handheld case, but completely fails for bodycams.
However, even in one of the handheld sequences the scale error for
SplineFusion is a magnitude larger than that of SEW.
It is unsurprising that the SplineFusion settings work on the
{\bf Handheld 2} dataset, as the motions in this dataset are relatively smooth
compared to the others.

\begin{figure}
  \begin{center}
    \includegraphics{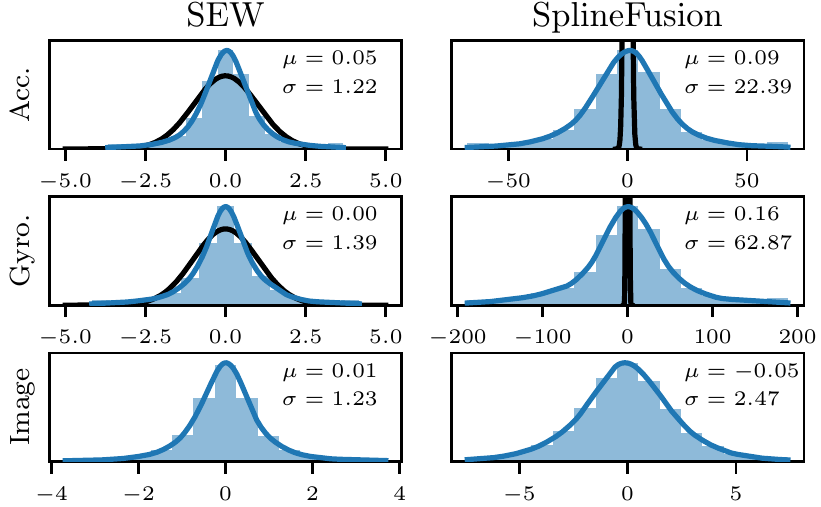}
    \caption{Blue: Residual distribution after convergence for the
      {\bf Handheld 2} dataset. 
    Black: PDF of the standard normal distribution $\mathcal{N}(0, 1)$ that is expected
    for the IMU residuals.}
    \label{fig:exp_residual_apple3}
  \end{center}
\end{figure}

In figure \ref{fig:exp_residual_apple3} we plot the residual
distributions after convergence on the {\bf Handheld 2} dataset.
Ideally the IMU residuals should be normally distributed with mean $\mu=0$ and standard deviation $\sigma=1$.
The image residuals could contain outliers, and are also mapped through the robust error norm $\phi$,
and are thus not necessarily normally distributed.
It is clear that SEW produces residuals which are close to standardized,
while the inverse noise covariance weighting of SplineFusion does not.

\begin{table}
  \begin{center}
    {\small \begin{tabular}{l|rr|rr}
\toprule
{} & \multicolumn{2}{c|}{EPE [m]} & \multicolumn{2}{c}{$e_\text{scale}$ [\%]} \\
 &  SEW &   SplineFusion &         SEW &     SplineFusion \\
\midrule
Bodycam 1  & {\bf 0.22} & 37.95 &        {\bf 3.4\%} & 124.2\% \\
Bodycam 2  & {\bf 0.40} & 27.86 &        {\bf 8.7\%} & 423.1\% \\
Handheld 1 & {\bf 0.30} &  0.52 &        {\bf 1.4\%} &  19.0\% \\
Handheld 2 & {\bf 0.56} &  0.94 &        2.0\% &   {\bf 1.7\%} \\
\bottomrule
\end{tabular}
}
  \end{center}
  \caption{Comparison of SEW and SplineFusion}
  \label{tab:sf_comp}
\end{table}

%------------------------------------------------------------------------
\section{Conclusions and future work}

We have introduced a method that balances residuals from different
modalities. In the experiments we applied the method to
continuous-time structure from motion, using measurements from
KLT-tracking and an IMU. This simple setup was used to highlight the
advantages with the proposed error weighting scheme. In order to
further improve the robustness in this particular application, one
could incorporate, by now classical ideas from SfM, such as
invariant features, and loop-closure detection.

The proposed spline error weighting uses empirical spectra in the
different measurement modalities. Such spectra are however often
characteristic of specific types of motion, \eg handheld and bodycam
sequences have characteristic within group 
spectra, and these could be learned and applied directly to new
sequences of the same type. Such learned characteristic spectra could
be useful as {\it a priori} information when adapting spline error
weighting to do on-line visual-inertial fusion.

Another potential improvement is to derive a better residual error
prediction for the accelerometer, that properly accounts for the
interaction with the sensor orientation changes. The quality
experiment in figure \ref{fig:exp_quality} revealed a consistent
underestimation of the approximation quality, and a better prediction
should remedy this.

We plan to release our spline error weighting framework for
visual-inertial fusion under an open source license.
\\
{\bf Acknowledgements:} This work was funded by the Swedish Research
Council through projects LCMM (2014-5928) and and EMC2
(2014-6227). The authors also thank Andreas Robinson for IMU logger hardware design.

{\small
\bibliographystyle{ieee}
\bibliography{local}

\begin{thebibliography}{10}\itemsep=-1pt

\bibitem{anderson15}
S.~Anderson and T.~D. Barfoot.
\newblock Full {STEAM} ahead: Exactly sparse gaussian process regression for
  batch continuous-time trajectory estimation on {SE}(3).
\newblock In {\em {IEEE} International Conference on Intelligent Robots and
  Systems ({IROS15})}, 2015.

\bibitem{anderson14}
S.~Anderson, F.~Dellaert, and T.~D. Barfoot.
\newblock A hierarchical wavelet decomposition for continuous-time {SLAM}.
\newblock In {\em {IEEE} International Conference on Robotics and Automation
  ({ICRA14})}, 2014.

\bibitem{anderson15b}
S.~Anderson, K.~MacTavish, and T.~D. Barfoot.
\newblock Relative continuous-time slam.
\newblock {\em The International Journal of Robotics Research},
  34(12):1453--1479, 2015.

\bibitem{bailey06}
T.~Bailey and H.~Durrant-Whyte.
\newblock {Simultaneous localization and mapping (SLAM): part II}.
\newblock {\em IEEE Robotics {\&} Automation Magazine}, 13(3):108--117, sep
  2006.

\bibitem{bouguet00}
J.-Y. Bouguet.
\newblock Pyramidal implementation of the lucas kanade feature tracker.
\newblock {\em Intel Corporation, Microprocessor Research Labs}, 2000.

\bibitem{brent73}
R.~P. Brent.
\newblock {\em Algorithms for Minimization without Derivatives}, chapter
  Chapter 4: An Algorithm with Guaranteed Convergence for Finding a Zero of a
  Function.
\newblock Englewood Cliffs, NJ: Prentice-Hall, 1973.

\bibitem{devernay01}
F.~Devernay and O.~Faugeras.
\newblock Straight lines have to be straight: Automatic calibration and removal
  of distortion from scenes of structured environments.
\newblock {\em Machine Vision and Applications}, 13(1):14--24, 2001.

\bibitem{forster15}
C.~Forster, L.~Carlone, F.~Dellaert, and D.~Scaramuzza.
\newblock {IMU} preintegration on manifold for efficient visual-inertial
  maximum-a-posteriori estimation.
\newblock In {\em Robotics: Science and Systems ({RSS'15})}, Rome, Italy, July
  2015.

\bibitem{furgale12}
P.~Furgale, T.~D. Barfoot, and G.~Sibley.
\newblock Continuous-time batch estimation using temporal basis functions.
\newblock In {\em {IEEE} International Conference on Robotics and Automation
  ({ICRA12})}, 2012.

\bibitem{furgale15}
P.~Furgale, C.~H. Tong, T.~D. Barfoot, and G.~Sibley.
\newblock Continuous-time batch trajectory estimation using temporal basis
  functions.
\newblock {\em International Journal of Robotics Research}, 2015.

\bibitem{gauglitz2011}
S.~Gauglitz, L.~Foschini, M.~Turk, and T.~Hollerer.
\newblock {Efficiently selecting spatially distributed keypoints for visual
  tracking}.
\newblock In {\em 18th {IEEE} International Conference on Image Processing},
  2011.

\bibitem{hanning11}
G.~Hanning, N.~Forsl\"ow, P.-E. Forss\'en, E.~Ringaby, D.~T\"ornqvist, and
  J.~Callmer.
\newblock Stabilizing cell phone video using inertial measurement sensors.
\newblock In {\em The Second {IEEE} International Workshop on Mobile Vision},
  2011.

\bibitem{hedborg12}
J.~Hedborg, P.-E. Forss\'en, M.~Felsberg, and E.~Ringaby.
\newblock Rolling shutter bundle adjustment.
\newblock In {\em {IEEE} Conference on Computer Vision and Pattern
  Recognition}, June 2012.

\bibitem{jia13}
C.~Jia and B.~L. Evans.
\newblock Online calibration and synchronization of cellphone camera and
  gyroscope.
\newblock In {\em {IEEE} Global Conference on Signal and Information Processing
  ({GlobalSIP})}, December 2013.

\bibitem{kopf14}
J.~Kopf, M.~F. Cohen, and R.~Szeliski.
\newblock First-person hyper-lapse videos.
\newblock In {\em SIGGRAPH Conference Proceedings}, 2014.

\bibitem{li13}
M.~Li, B.~H. Kim, and A.~I. Mourikis.
\newblock Real-time motion tracking on a cellphone using inertial sensing and a
  rolling-shutter camera.
\newblock In {\em {IEEE} International Conference on Robotics and Automation
  {ICRA'13}}, 2013.

\bibitem{lovegrove13}
S.~Lovegrove, A.~Patron-Perez, and G.~Sibley.
\newblock Spline fusion: A continuous-time representation for visual-inertial
  fusion with application to rolling shutter cameras.
\newblock In {\em British Machine Vision Conference ({BMVC})}. {BMVA},
  September 2013.

\bibitem{mihajlovic1999}
Z.~Mihajlovic, A.~Goluban, and M.~Zagar.
\newblock {Frequency Domain Analysis of B-Spline Interpolation}.
\newblock {\em ISIE'99 - Bled, Slovenia}, pages 193--198, 1999.

\bibitem{oth13}
L.~Oth, P.~Furgale, L.~Kneip, and R.~Siegwart.
\newblock Rolling shutter camera calibration.
\newblock In {\em {IEEE} Conference on Computer Vision and Pattern Recognition
  ({CVPR13})}, pages 1360--1367, Portland, Oregon, June 2013.

\bibitem{ovren15}
H.~Ovr\'en and P.-E. Forss\'en.
\newblock Gyroscope-based video stabilisation with auto-calibration.
\newblock In {\em {IEEE} International Conference on Robotics and Automation
  {ICRA'15}}, 2015.

\bibitem{patron-perez15}
A.~Patron-Perez, S.~Lovegrove, and G.~Sibley.
\newblock A spline-based trajectory representation for sensor fusion and
  rolling shutter cameras.
\newblock {\em International Journal on Computer Vision}, 113(3):208--219,
  2015.

\bibitem{rosten10}
E.~Rosten, R.~Porter, and T.~Drummond.
\newblock Faster and better: A machine learning approach to corner detection.
\newblock {\em IEEE Trans. Pattern Anal. Mach. Intell.}, 32(1), Jan. 2010.

\bibitem{sibley09}
G.~Sibley, C.~Mei, and I.~Reid.
\newblock Adaptive relative bundle adjustment.
\newblock In {\em Robotics: Science and Systems ({RSS'09})}, 2009.

\bibitem{triggs00}
B.~Triggs, P.~Mclauchlan, R.~Hartley, and A.~Fitzgibbon.
\newblock Bundle adjustment – a modern synthesis.
\newblock In {\em Vision Algorithms: Theory and Practice, LNCS}, pages
  298--375. Springer Verlag, 2000.

\bibitem{unser99}
M.~Unser.
\newblock Splines -- a perfect fit for signal and image processing.
\newblock {\em {IEEE} Signal Processing Magazine}, 16(6):22--38, 1999.

\bibitem{unser1993}
M.~Unser, A.~Aldroubi, and M.~Eden.
\newblock {B-spline signal processing. II. Efficiency design and applications}.
\newblock {\em IEEE Transactions on Signal Processing}, 41(2):834--848, 1993.

\end{thebibliography}
}

\end{document}